\documentclass[letterpaper, 10 pt, conference]{ieeeconf}  % Comment this line out if you need a4paper
\usepackage[utf8]{inputenc}
\usepackage[english]{babel}

\IEEEoverridecommandlockouts                              % This command is only needed if
\usepackage{graphicx}

\usepackage{xcolor}

\usepackage{subfig} % subfloat
\usepackage[export]{adjustbox}

\usepackage{amsmath,amsfonts,amssymb,bm}
\usepackage{nicefrac}

\usepackage{siunitx}
\DeclareSIUnit \parsec {pc}
\DeclareSIUnit \electronvolt {eV}
\DeclareSIUnit \pixel {px}
\DeclareSIUnit \arcmin {arcmin}
\DeclareSIUnit \erg {erg}
\DeclareSIUnit \joul {J}

\usepackage{tikz}
\usepackage{pgfplots}
\pgfplotsset{compat=1.14}
\usetikzlibrary{backgrounds,arrows,automata,shapes,positioning,calc,through,spy}
\pgfdeclarelayer{background}
\pgfdeclarelayer{foreground}
\pgfsetlayers{background,main,foreground}

\usepackage[]{hyperref}

\usepackage{isotope}

\tikzset{
  imglabel/.style={
    rectangle,
    inner sep=2pt,
    rounded corners=.1em,
    text=black,
    minimum height=1em,
    text centered,
    fill=white,
    fill opacity=1.0,
    text opacity=1,
    anchor=south west,
    draw=black,
  },
  imgletter/.style={
    rectangle,
    inner sep=2pt,
    text=black,
    minimum height=1em,
    text centered,
    fill=white,
    fill opacity=0.7,
    text opacity=1,
    anchor=south west,
  },
  double arrow/.style args={#1 colored by #2 and #3}{
    -stealth,line width=#1,#2, % first arrow
    postaction={draw,-stealth,#3,line width=(#1)/3,
    shorten <=(#1)/3,shorten >=2*(#1)/3}, % second arrow
  }
}

\usepackage[nolist,nohyperlinks]{acronym}
\acrodef{GPS}[GPS]{Global Positioning System}
\acrodef{SLAM}[SLAM]{Simultaneous Localization And Mapping}
\acrodef{SLAMs}[SLAMs]{Simultaneous Localization And Mapping systems}
\acrodef{GPS}[GPS]{Global Positioning System}
\acrodef{RTK}[RTK]{Real-time Kinematics}
\acrodef{GNSS}[GNSS]{Global Navigation Satellite System}
\acrodef{ROS}[ROS]{Robot Operating System}
\acrodef{API}[API]{Application Programming Interface}
\acrodef{UAV}[UAV]{Unmanned Aerial Vehicle}
\acrodef{MAV}[MAV]{Micro Aerial Vehicle}
\acrodef{UGV}[UGV]{Unmanned Ground Vehicle}
\acrodef{UV}[UV]{Ultra-Violet}
\acrodef{LED}[LED]{Light-emitting Diode}
\acrodef{MBZIRC}[MBZIRC]{Mohamed Bin Zayed International Robotics Challenge}
\acrodef{DARPA}[DARPA]{Defense Advanced Research Projects Agency}
\acrodef{IMU}[IMU]{Inertial Measurement Unit}
\acrodef{LTI}[LTI]{Linear time-invariant}
\acrodef{MPC}[MPC]{Model Predictive Control}
\acrodef{UVDAR}[UVDAR]{Ultra-Violet Direction And Ranging}
\acrodef{DOF}[DOF]{degree-of-freedom}
\acrodef{DOFs}[DOFs]{degrees-of-freedom}
\acrodef{LiDAR}[LiDAR]{Light Detection and Ranging}
\acrodef{ESC}[ESC]{Electronic Speed Controller}
\acrodef{LKF}[LKF]{Linear Kalman Filter}
\acrodef{UKF}[UKF]{Unscented Kalman Filter}
\acrodef{EKF}[EKF]{Extended Kalman Filter}
\acrodef{RAS}[RAS]{Robotics and Automation Society}
\acrodef{IEEE}[IEEE]{Institute of Electrical and Electronics Engineers}
\acrodef{MRS}[MRS]{Multi-robot Systems Group}
\acrodef{FOV}[FOV]{Field of View}
\acrodef{CdTe}[CdTe]{Cadmium Telluride}
\acrodef{FDNPP}[FDNPP]{Fukushima Daiichi Nuclear Power Plant}

\newcommand{\m}[1]{\ensuremath{\mathbf{#1}}}
\newcommand\numberthis{\addtocounter{equation}{1}\tag{\theequation}}

\newcommand{\reffig}[1]{Fig.~\ref{#1}}

\newcommand{\minus}{\scalebox{0.75}[1.0]{$-$}}

\title{\LARGE \bf
Gamma Radiation Source Localization for Micro Aerial Vehicles with~a~Miniature Single-Detector Compton Event Camera
}

\author{Tomas Baca$^{1}$, Petr Stibinger$^{1}$, Daniela Doubravova$^{2}$, Daniel Turecek$^{2}$, \\Jaroslav Solc$^{3}$, and Jan Rusnak$^{3}$, Martin Saska$^{1}$, Jan Jakubek$^{2}$% <-this % stops a space
\thanks{$^{1}$Authors are with Czech Technical University in Prague, Technicka 2, 160 00, Prague.
{\tt\small {tomas.baca}@fel.cvut.cz}}%
\thanks{$^{2}$Authors are with Advacam s.r.o., U Pergamenky 12, 170 00, Prague.}%
\thanks{$^{3}$Authors are with Czech Metrology Institute, Okruzni 31, 638 00, Brno.}%
}

\begin{document}

\maketitle
\thispagestyle{empty}
\pagestyle{empty}

%%{ Abstract

\begin{abstract}
  A novel method for localization and estimation of compact sources of gamma radiation for \acp{MAV} is presented in this paper.
  The method is developed for a novel single-detector Compton camera, developed by the authors.
  The detector is extremely small and weighs only 40\,g, which opens the possibility for use on sub-1\,kg class of drones.
  The Compton camera uses the MiniPIX TPX3 CdTe event camera to measure Compton scattering products of incoming high-energy gamma photons.
  The 3D position and the sub-nanosecond time delay of the measured scattering products are used to reconstruct sets of possible directions to the source.
  The proposed method utilizes a filter for fusing the measurements and estimating the radiation source position during the flight.
  The computations are executed in real-time onboard and allow integration of the detector info into a fully-autonomous system.
  Moreover, the real-time nature of the estimator potentially allows estimating states of a moving radiation source.
  The proposed method was validated in simulations and demonstrated in a real-world experiment with a Cs137 radiation source.
  The approach can localize a gamma source without estimating the gradient or contours of radiation intensity, which opens possibilities for localizing sources in a cluttered and urban environment.
\end{abstract}

%%}

%%{ Introduction

\section{INTRODUCTION}

Localization of ionizing radiation sources has recently become the domain of autonomous mobile robots.
Following the 2011 disaster at the \ac{FDNPP}, considerable efforts went into minimizing the exposure of human workers by sending \acp{UGV} into potentially hazardous areas.
Mapping a radiation distribution within an area is traditionally done with a dosimeter (measures intensity of radiation) while scanning the whole area.
On the other hand, the search for compact sources is better done with sensors capable of estimating the direction to the source.
The direction estimation is traditionally achieved by pixel radiation detectors (cameras) with sensor attachments, such as pinhole optics, coded apertures, and collimators \cite{baca2019timepix}.
In recent years, Compton cameras made an addition to the list of ways to estimate a set of directions to the source, however, without an external attachment to the sensor.
A \emph{traditional} Compton camera provides the direction to a source by measuring products of a Compton scattering reaction within two detectors (scatter and absorber).
Information on the products is used to calculate a set of possible directions to the source in the form of a cone surface in 3D.

%%{ Fig: drone and timepix

\begin{figure}
  \centering
  \subfloat {
    \includegraphics[width=0.195\textwidth]{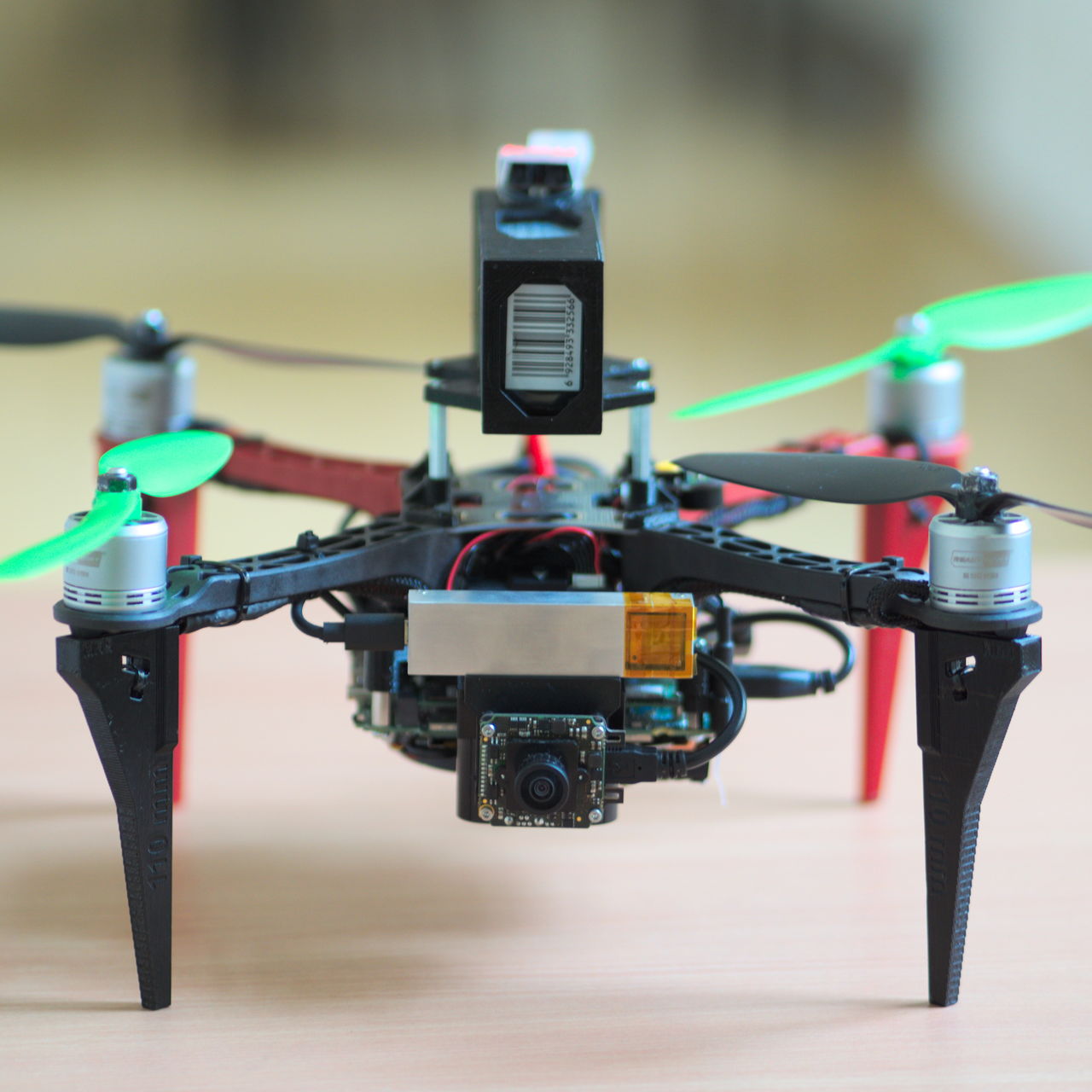}
  }%
  \subfloat {
    \includegraphics[width=0.25\textwidth]{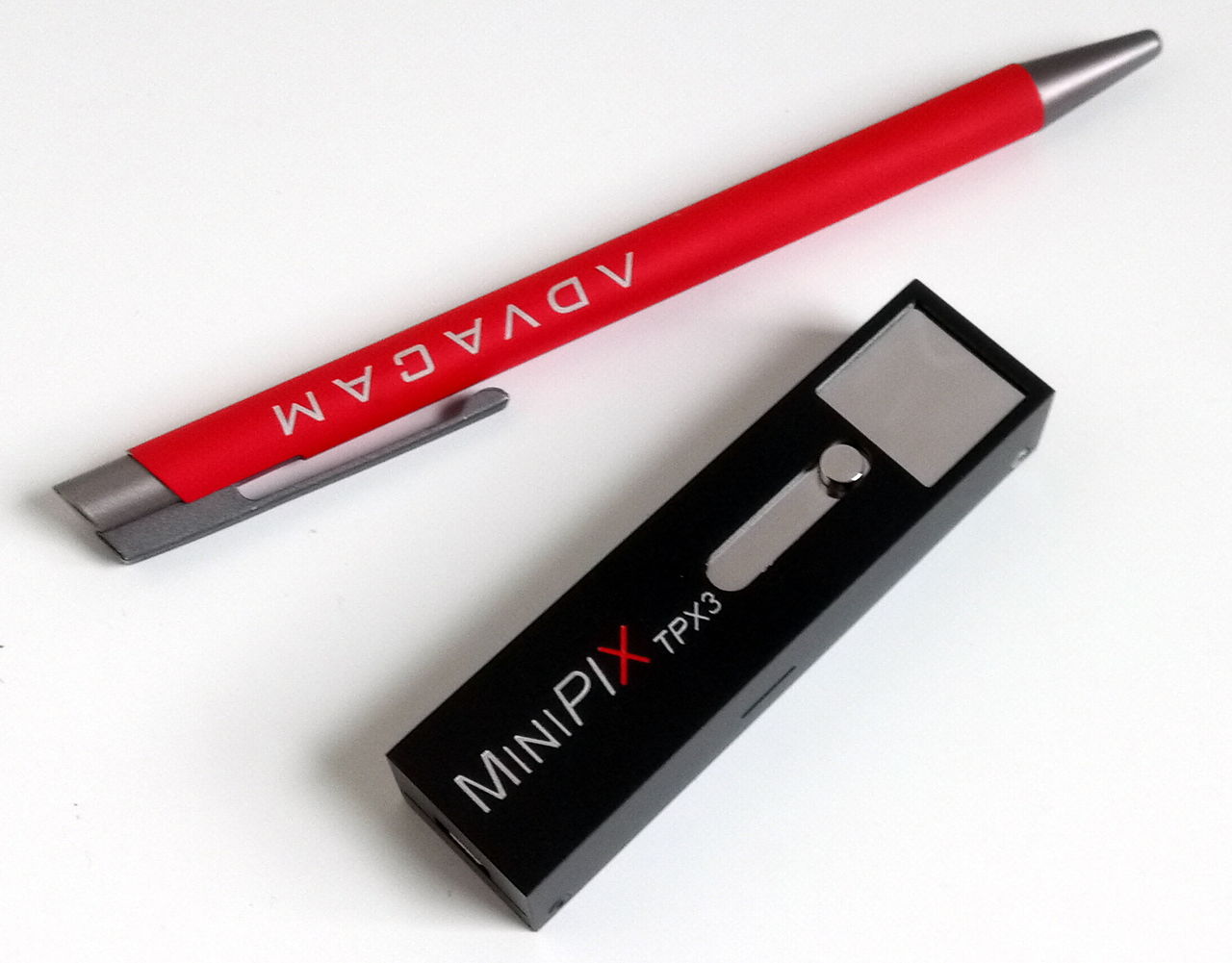}
  }%
  \hfill%
  \caption{A miniature \ac{UAV}, one of our experimental platforms, equipped with the MiniPIX TPX3 Compton camera.}
  \label{fig:timepix}
\end{figure}

%%}

We aim to bring the Compton camera to the world of \acp{MAV} --- sub-\SI{1}{\kilo\gram} \acp{UAV}.
With current state-of-the-art technologies, \acp{MAV} can fly autonomously in cluttered environments, such as urban and indoor environments and forests.
An \ac{MAV} can pursue an informed exploration of the environment towards the estimated position of the source if such a real-time estimate is available.

%%{ Sub: Related works

\subsection{Related work}

% In the field of radiation sensing, unmanned robotic vehicles offer several advantages over conventional handheld detectors or piloted aircraft.
% These advantages can be exploited in a wide variety of applications.

% Following the 2011 disaster at the \ac{FDNPP}, considerable amounts of radioactive material have been released into the plant area.

Several \acp{UGV} have been deployed directly inside the damaged reactor buildings of \ac{FDNPP} under remote control.
Various radiation detection methods have been tested inside the power plant, including a coded aperture scintillator \cite{ohno2011robotic}, a semiconductor digital dosimeter \cite{nagatani2013emergency}, a Compton event camera composed of two scintillators \cite{sato2019radiation}, and a time-of-flight gamma camera \cite{kinoshita2014development}.
Ground-based robots offer higher payload capacity and the ability to carry heavier sensory equipment than most aerial vehicles.
On the other hand, these robots tend to be relatively bulky, struggle to navigate the cluttered corridors and staircases inside the damaged buildings, and generally move slower than a multirotor aircraft.

\acp{UAV} have been utilized to map the spread of the radioactive material outside of the power plant.
These range from large aircraft weighing more than \SI{90}{\kilogram} equipped with heavy scintillation detectors \cite{sanada2015aerial, towler2012radiation, jiang2016prototype}
to compact multi-rotors suitable for flying along a pre-defined trajectory close to the ground \cite{macfarlane2014lightweight, christie2017radiation, martin20163d}.
Outside of Japan, several projects have employed \acp{UAV} for radiation intensity mapping around uranium ore mines \cite{salek2018mapping, keatley2018source, martin2015use}.

In \cite{han2013lowcost}, multiple fixed-wing \acp{UAV} equipped with miniature scintillators are used for contour analysis of an irradiated area.
Trajectory planning and data processing are performed offline, contrary to our approach, which estimates the position of the source in real-time during the flight.
In \cite{newaz2016uav}, the contour analysis is tackled using a single multirotor \ac{UAV}. The contour analysis uses a Gaussian mixture model to estimate multiple radiation sources' positions with overlapping intensity fields.
The projects mentioned above utilize unmanned vehicles to deliver a radiation detector into a hazardous environment.
However, the approaches do not respond to measured data in real-time and thus do not exploit the mobility of \acp{UAV} to improve the measurement.

Active path-planning driven by the onboard measurements has been shown in \cite{towler2012radiation} for an outdoor environment and in \cite{mascarich2018radiation} for a GPS-denied indoor environment.
Both of these works rely on a scintillator detector to estimate the radiation intensity in the \ac{UAV}'s current position.
As a result, the employed aerial platforms have to be large with a payload capacity exceeding \SI{2}{\kilogram}.
As in \cite{towler2012radiation}, the aerial platform is a \SI{90}{\kilogram} unmanned aerial helicopter, which significantly limits its deployment conditions due to personal safety and considerable minimum distance to obstacles in the environment.
Similarly, scintillating detectors were used in \cite{mascarich2018radiation} on a mobile ground robot to map a cluttered environment and to search for radiation sources.

The lack of lightweight radiation detectors with immediate readout capability severely limits the application potential of aerial dosimetry.
However, a new robotic methodology for motion planning and exploration needs to be developed to accommodate the proposed measurement system's specifics.
The proposed system is intended for outdoor and indoor environments, prompting a very compact and self-sufficient vehicle.
We build upon our previous experience with indoor \ac{UAV} deployment \cite{baca2021mrs}, where such radiation localization system could be used during exploration of mine shafts and caves \cite{roucek2019darpa, petrlik2020robust}, in interiors of large buildings \cite{kratky2020autonomous, saska2020formation}, in forests \cite{petracek2020bioinspired} and in nuclear power plants \cite{kratky2020autonomous2}.

%%}

%%{ Tab: nomenclature

\begin{table*}
  \scriptsize
  \centering
  \noindent\rule{\textwidth}{0.5pt}
  \begin{tabular}{lll}
    $\mathbf{x}$, $\bm{\alpha}$ & vector, pseudo-vector, or tuple\\
    $\mathbf{\hat{x}}$, $\bm{\hat{\omega}}$& unit vector or unit pseudo-vector\\
    $\mathbf{\hat{e}}_1, \mathbf{\hat{e}}_2, \mathbf{\hat{e}}_3$ & elements of the \emph{standard basis} \\
    $\mathbf{X}, \bm{\Omega}$ & matrix \\
    $x = \mathbf{a}^\intercal\mathbf{b}$ & inner product of $\mathbf{a}$, $\mathbf{b}$ $\in \mathbb{R}^n$\\
    $\mathbf{x} = \mathbf{a}\times\mathbf{b}$ & cross product of $\mathbf{a}$, $\mathbf{b}$ $\in \mathbb{R}^3$\\
  \end{tabular}%
  \begin{tabular}{lll}
    $\mathbf{x} = \mathbf{a}\circ\mathbf{b}$ & element-wise product of $\mathbf{a}$, $\mathbf{b}$ $\in \mathbb{R}^n$ \\
    $x_{[n]}$ & $x$ at the sample $n$ \\
    $\mathbf{A}, \mathbf{x}, \mathbf{\Omega}$ & LKF system matrix, state vector, covariance\\
    $\Delta t$ & time difference interval, [\si{\second}] \\
    $\mathbf{x}_\mathcal{B}, \mathbf{x}_\mathcal{W}, \mathbf{x}_\mathcal{C}$ & $\mathbf{x}$ in body-fixed, world, and camera frame \\
    $\angle\left(\mathbf{a}, \mathbf{b}\right)$ & signed angle between vectors (right-hand rule) \\
  \end{tabular}
  \noindent\rule{\textwidth}{0.5pt}
  \caption{Mathematical notation, nomenclature, and notable symbols.}
  \label{tab:nomenclature}
\end{table*}

%%}

%%{ Sub: Contributions

\subsection{Contributions}

We present a novel method for gamma source radiation localization using a single-detector Compton camera \cite{turecek2020single}.
The method utilizes an onboard position estimator of a \ac{UAV} to estimate a real-time hypothesis of the radiation source position.
The proposed method is executed in real-time such that the results can be used to drive an autonomous search algorithm.
Unlike any existing solution, the system is deployable onboard \aclp{MAV} thanks to the small form factor of the utilized Compton camera detector.
The proposed approach is intended as an addition to an intensity-based estimation approach and the proposed method can, on its own, maintain a feasible hypothesis of a moving radiation source.
The used MiniPIX TPX3 detector with \ac{CdTe} sensor can also detect $\beta$ and heavy-ion radiation and, therefore, can be used without changes to estimate gradients and contours of radiation intensity.
Even more, particle tracks can be classified \cite{baca2019timepix} to deduce the nature of the radiation source.

%%}

%%}

%%{ Compton camera

\section{TIMEPIX3 COMPTON CAMERA}

The process of recording and imaging ionizing gamma radiation is very different from the near-visible light spectrum photography \cite{baca2019timepix}.
Refractive optical lenses cannot be made for such energies, and reflective optics (Kirkpatrick-Baez, Walter, Lobster-eye) are limited to energies not exceeding \SI{30}{\kilo\electronvolt} by the reflectivity of the used mirror materials such as gold and silicon.
Pinhole gamma cameras are a viable option.
However, their \ac{FOV} is narrow, and the aperture shield absorbs a valuable portion of the incoming particles.

\begin{figure}[!t]
  \centering
  \resizebox{0.25\textwidth}{!}{%
    \begin{tikzpicture}[thick]
  \tikzset{snake it/.style={decorate, decoration=snake}}
    \draw[draw=red,snake it] (0.0, 0.0) -- (2.0, 0.0);
    \node at (0.0, 0.3) {$\lambda$};

    \draw[draw=green,snake it] (2.0, 0.0) -- (3.414, 1.414);
    \node at (3.614, 1.414) {$\lambda'$};

    \draw[draw=black,dashed] (2.0, 0.0) -- (3.0, 0.0);

    \draw[->, draw=black] (2.0, 0.0) -- (3.0, -0.5);
    \node at (3.2, -0.5) {$e^{\minus}$};

    \draw[black, dashed] (3,0) arc (0:40:1);
    \node at (2.7, 0.25) {$\Theta$};

    \fill[circle,shading=ball,outer color=gray!30] (2.0,0) circle (5pt);
\end{tikzpicture}
  }
  \caption{
    Compton scattering: an incoming gamma photon $\lambda$ changes its momentum as it scatters off a scattering center by a scattering angle $\Theta$.
    The decrease in momentum creates an electron.
  }
  \label{fig:compton_scattering}
\end{figure}
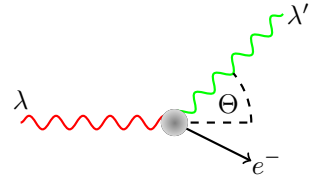

Compton cameras use the Compton scattering effect (see \reffig{fig:compton_scattering}) to reconstruct a projection of a radioactive source \cite{turecek2018compton}.
Typically, a Compton camera consists of two sensors.
The first detector, a \emph{scatterer}, witnesses the Compton scattering effect in its sensor by capturing the bi-product of the reaction --- the electron $e^{-}$.
The electron is captured immediately upon its creation.
The second sensor is placed behind the first and is dedicated to capturing the scattered photon $\lambda'$ through the photo-electric effect.
When both detectors detect the scattering products, they simultaneously measure the energy and the 3D position of the particles.
According to Compton,~1923~\cite{compton1923quantum}, the relation of energies of the particles to the scattering angle $\Theta$ is modelled as
\begin{equation}
  E_{\lambda'} = \frac{E_{\lambda}}{1 + \left(E_{\lambda} / \left(m_e c^2\right)\right)\left(1 - \cos\Theta\right)},
  \label{eq:compton_formulae}
\end{equation}
where $E_{\lambda}, E_{\lambda'}$ are the energies of the incoming and scattered photon (\si{\joul}), $E_{e^{\minus}}$ is the energy of the electron (\si{\joul}), $m_e = 9.10938356\times10^{-31}\,$\si{\kilogram} is the electron rest mass, and $c = 299792458$\,\si{\meter\per\second} is the speed of light in vacuum.
Since Compton scattering is a radially symmetrical effect, the reconstruction of $\Theta$ yields a set of possible incoming directions that forms a cone surface.

%%{ Fig: Compton scattering

%%}

Contrary to recent works \cite{sato2019radiation, jiang2016prototype, terzioglu2018compton}, we employ only a single sensor to measure the scattering products \cite{turecek2020single}.
Using cutting-edge innovations in the field of particle imaging, we employ the Timepix3 detector \cite{poikela2014timepix3} as part of the MiniPIX TPX3 event camera with a thick \SI{2}{\milli\meter} \ac{CdTe} sensor \footnote{\url{https://advacam.com/camera/minipix-tpx3}}.
The thick semiconductor sensor is capable of causing the scattering effects, as well as capturing the products.
Thanks to the detector's event camera operation mode, both bi-products are timed with sub-nanosecond accuracy, and the depth of the event within the sensor is calculated.
In contrast with traditional frame-based cameras, event-driven cameras output a continuous stream of data generated by the active (hit) pixels.
A similar trend emerged in the visible-light camera field \cite{kueng2016low}.
Timepix3 promises a similar impact in the field of aerial radiation detection, as its essential properties are perfectly suited for dynamically flying \acp{MAV}.

The coincidence detection window is set to \SI{86}{\nano\second}, which is the maximum time difference of two coinciding particles being measured at the opposite side of the \SI{2}{\milli\meter} \ac{CdTe} at \SI{450}{\volt}.
The detection window is specific to this particular sensor and should be adjusted for different thicknesses, materials, and bias voltages.

\subsection{Reconstructing the set of directions to the source}

%%{ Fig: 3D Compton scattering

\begin{figure}[!b]
  \centering
  \includegraphics[width=0.30\textwidth]{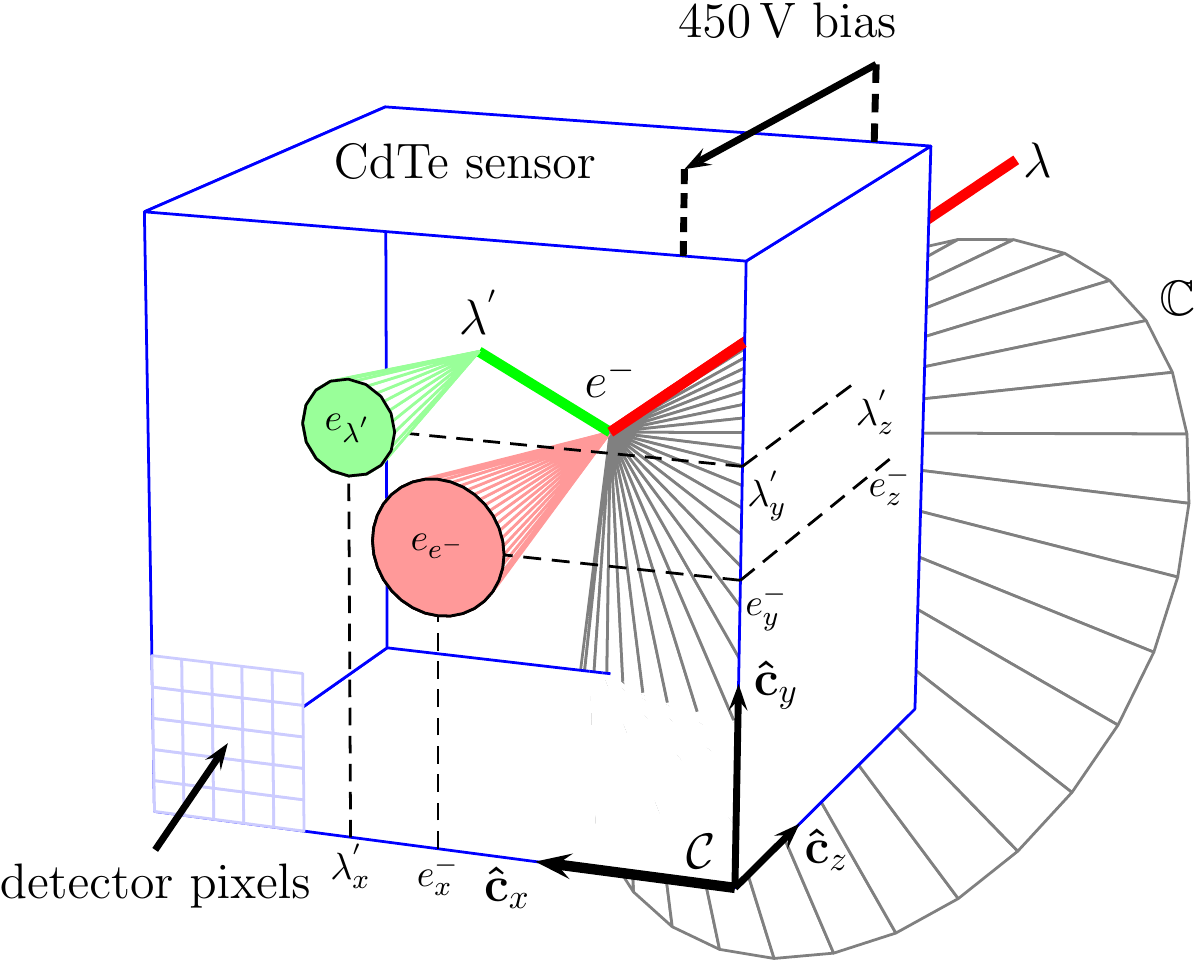}
  \caption{
    An illustration of the Compton scattering effect within the \ac{CdTe} sensor on top of the Timepix3 detector.
    The ionized electron clouds are produced by absorption $\lambda'$ and $e^-$ and are gathered towards the detector pixels by $\SI{450}{\volt}$ bias potential with a speed of $\approx$\,\SI{23}{\micro\meter\per\nano\second}.
  }
  \label{fig:3d_cone_reconstruction}
\end{figure}

%%}

\begin{figure}[!b]
  \centering
  \includegraphics[width=0.35\textwidth]{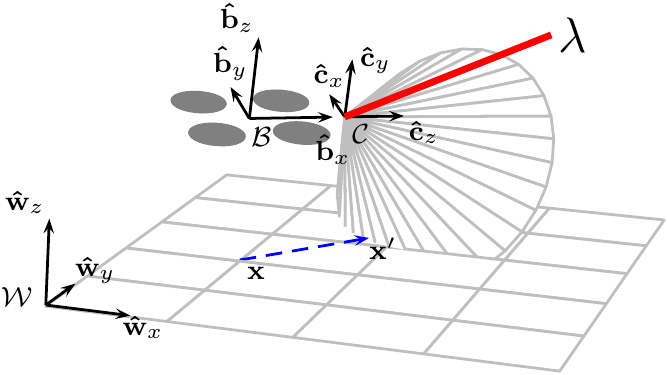}
  \caption{
    Compton camera coordinate frame $\mathcal{C}$ is attached to the \ac{UAV} body frame $\mathcal{B}$.
    The world coordinate frame $\mathcal{W}$ is static. Transformations between the frames are provided by the \ac{UAV} control system \cite{baca2021mrs} through \ac{ROS}.
  }
  \label{fig:sketch_coordinate_frames}
\end{figure}

\begin{figure}
  \centering
  \begin{tikzpicture}
    \node[anchor=south west,inner sep=0] (a) at (0,0) {
        \begin{tabular}{cc}
            \fbox{\includegraphics[width=0.20\textwidth]{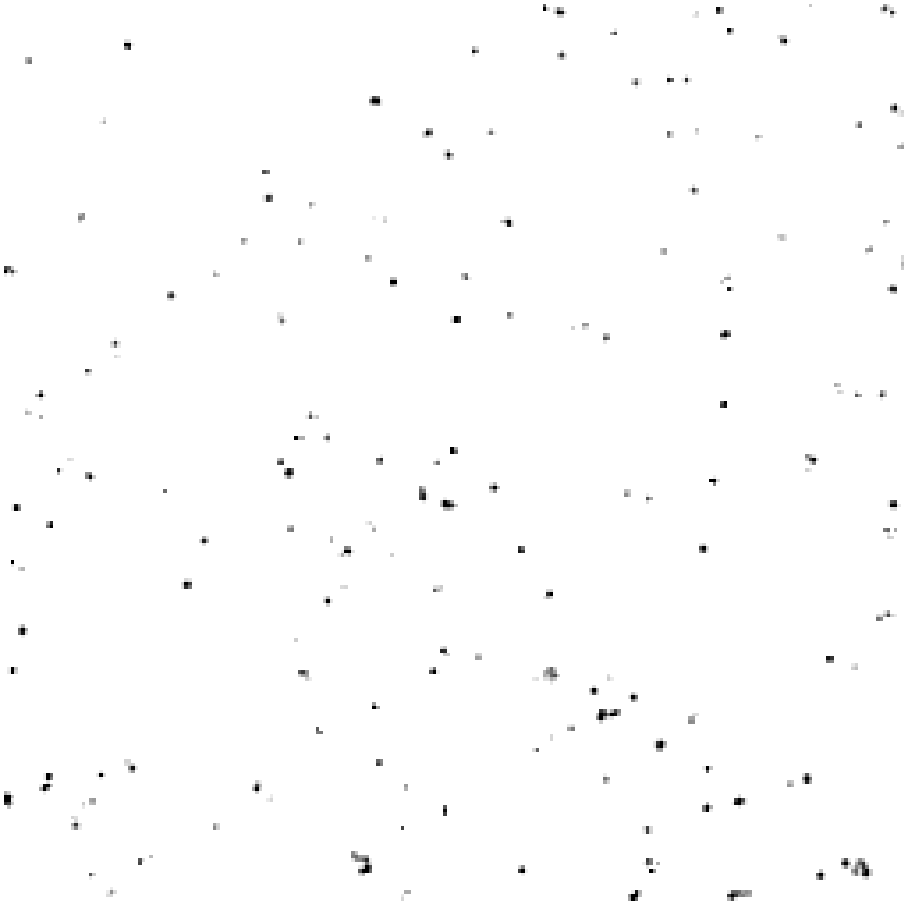}}
          &
            \fbox{\includegraphics[width=0.20\textwidth]{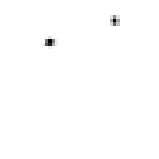}}
        \end{tabular}
      };

    %%{ labels

    \begin{scope}[x={(a.south east)},y={(a.north west)}]

      \draw (0.37, 0.11) rectangle (0.43, 0.20);
      \draw[black] (0.43, 0.11) -- (0.525, 0.005);
      \draw[black] (0.43, 0.20) -- (0.525, 0.995);

      \draw (0.66,0.80) node [text=black] {\small photon track};
      \draw (0.83,0.95) node [text=black] {\small electron track};

      \draw (0.75,0.33) node [text=black] {
        \begin{tabular}{rl}
          \small $\lambda'_e$ = &\SI{394.22}{\kilo\electronvolt}\\
          \small $e^{\minus}_e$ = &\SI{315.70}{\kilo\electronvolt}\\
          \small $\Delta t$ = &\SI{20.31}{\nano\second}\\
          \small $\Delta z$ = &\SI{0.47}{\milli\meter}\\
          \hline
          \small $\Theta$ = &\SI{1.13}{\radian}\\
        \end{tabular}
      };

      %%{ grid

      % useful grid to help you find coordinates for plotting the overlay
      % \draw[black, xstep=.1, ystep=.1] (0,0) grid (1,1);
      % \foreach \i in {0,0.1,0.2,0.3,0.4,0.5,0.6,0.7,0.8,0.9,1} {
      %   \node[align=center] at (\i, -0.05) {\i};
      %   \node[align=center] at (\i, 1.05) {\i};
      %   \node[align=center] at (-0.05, \i) {\i};
      %   \node[align=center] at (1.05, \i) {\i};
      % }

      %%}

    \end{scope}

    %%}
  \end{tikzpicture}
  \caption{
    An example of a pair of Compton scattering products captured by the Timepix3 detector.
    The events' times are used together with the particles' energies to reconstruct the scattering angle, $\Theta$.
  }
  \label{fig:timepix_image}
\end{figure}
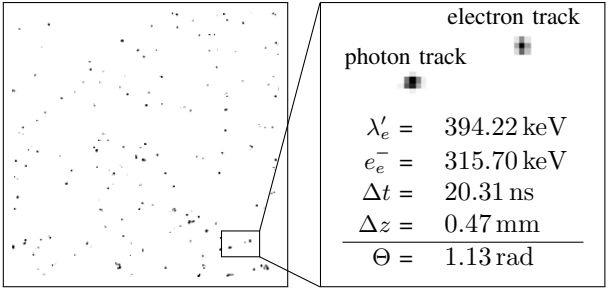

Table~\ref{tab:nomenclature} summarizes the mathematical notation used throughout this work and \reffig{fig:sketch_coordinate_frames} provides an illustration of the used coordinate systems.
Ionizing radiation separates electrons (ionization) from the \ac{CdTe} sensor material in the location of absorption.
The newly created electron cloud is accelerated by a \SI{450}{\volt} electric potential towards one facet of the \SI{2}{\milli\meter} \ac{CdTe} sensor where the charge is measured by individual pixels of the Timepix3 detector.
Figure \ref{fig:3d_cone_reconstruction} shows the charge-gathering process within the sensor.
The $[\lambda'_x, \lambda'_y]^\intercal_\mathcal{C}$ and $[e^{\minus}_x, e^{\minus}_y]^\intercal_\mathcal{C}$ coordinates of the events are obtained by calculating the centroids of the respective pixel tracks (see \reffig{fig:timepix_image}).
The z-coordinates $\lambda'_z$ and $e^{\minus}_z$ are unknown.
However, it is possible to estimate the relative difference along the $\mathbf{\hat{c}}_z$ using the measured time-difference of the events.
Although the absorption of $\lambda'$ and $e^{\minus}$ happens virtually at the same time\footnote{The speed of a gamma photon through \ac{CdTe} material is near the speed of light in vacuum and therefore can be neglected.}, the charge-gathering causes a delay between the measured electron clouds.
The relative z-distance is obtained as
\begin{equation}
  \Delta z = 2325.6\,(e^{\minus}_t - \lambda'_t),
\end{equation}
where $e^{\minus}_t$ and $\lambda'_t$ are the times of arrival of the electron clouds from the electron and photon, respectively, and $c_s$~=~\SI{23.256}{\micro\meter\per\nano\second}~=~\SI{2325.6}{\meter\per\second} is the charge-gathering speed through the \ac{CdTe} sensor with \SI{450}{\volt} bias\footnote{The charge gathering speed is obtained empirically by analyzing tracks of muons from a cosmic radiation.}.
The absence of a precise absolute z-axis coordinate of the cone origin within the camera frame is inconsequential.
It results only in a shift of the resulting cone surface, which can be neglected compared to the real-world distances involved.
Only the difference between the scattering products along $\mathbf{\hat{c}}_z$ is significant for estimating the cone parameters.
Therefore, coordinates of the electron and the secondary photon events in the camera frame $\mathcal{C}$ are considered as
\begin{align*}
  \mathbf{e}^{\minus}_{\mathcal{C}} &= [e^{\minus}_x, e^{\minus}_y, \Delta z]^\intercal,\numberthis\\
  \bm{\lambda}'_\mathcal{C} &= [\lambda'_x, \lambda'_x, 0]^\intercal.\numberthis
\end{align*}

%%{ Fig: Coordinate frames

%%}

In order to reconstruct the scattering angle $\Theta$, we convert the measured particle energies from (\si{\kilo\electronvolt}) to (\si{\joul}) as
\begin{align*}
  E_{e^{\minus}} &= e_{e^{\minus}}/6.242\times10^{18},\numberthis\\
  E_{\lambda} &= e_{\lambda}/6.242\times10^{18},\numberthis
\end{align*}
and calculate the scattering angle using (\ref{eq:compton_formulae}) as:
\begin{align*}
  \Theta = \cos^{-1} \underbrace{\left(1 + m_e c^2 \left(\frac{1}{E_{e^{-}} + E_\lambda} - \frac{1}{E_\lambda}\right)\right)}_{B},\\\text{for}\ \minus 1 < B < 1\numberthis
  \label{eq:theta_equation}
\end{align*}
Finally, the cone parameters are the cone origin $\mathbf{o}_{\mathcal{C}} = \mathbf{e}^{\minus}$, the direction $\mathbf{\hat{d}}_\mathcal{C} = \mathbf{e}^{\minus} - \bm{\lambda}'$, and the inner angle $\Theta$.

%%{ Fig: Timepix images

%%}

%%}

%%{ Radiation source state estimation

\section{RADIATION SOURCE STATE ESTIMATION}

Estimating the direction to the source is typically done under assumptions that do not apply in scenarios involving mobile robots.
The camera position is often fixed during the whole measurement, and all the data are processed at once after the measurement \cite{turecek2018compton, terzioglu2018compton}.
In other cases, the camera needs to be stationary to complete a single measurement set (tens of seconds) before it can be moved further \cite{sato2019radiation, jiang2016prototype}.
However, the estimation should be conducted continuously without stopping to utilize the advantages of multirotor helicopters fully.
Moreover, in all the previous works involving a mobile robot, the used camera was a heavy device, which required the use of a large \ac{UGV} or \ac{UAV}.
We aim to bring the Compton camera detector to the domain of \acp{MAV}.
With appropriate onboard localization and mapping systems, such an \ac{MAV} can automatically perform the search for an ionizing radiation source.
However, this requires the \ac{MAV} to automatically estimate the source's position onboard in real-time to incorporate newly acquired data on the fly.
Here we present our initial method developed for a single radiation source localization.

The proposed method relies on the \ac{LKF} to estimate a hypothesis and a covariance of the source's position.
However, the \ac{LKF} is not natively able to accept the measured cone surfaces as a vector of measurement.
To tackle this issue, we assume that the optimal correction would move the hypothesis orthogonally in the cone surface direction.
Such corrections behave idempotently under ideal conditions.
Applying such a correction requires a novel approach of using \ac{LKF} by defining an orthogonal projection to the cone surface.

%%{ Sub: Orthogonal projection to a cone surface

\subsection{Orthogonal projection to a cone surface}

\begin{figure}
  \centering
  \includegraphics[width=0.35\textwidth]{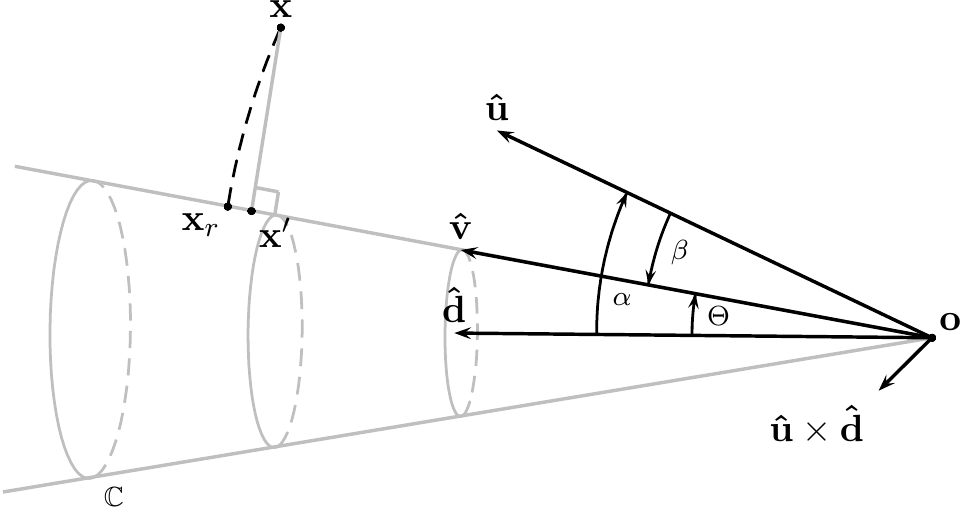}
  \caption{Orthogonal projection of $\mathbf{x}$ to the cone surface.}
  \label{fig:orthogonal_projection}
\end{figure}

Projecting the \ac{LKF} hypothesis $\mathbf{x}_\mathcal{W}$ orthogonally to the cone surface to obtain ${\mathbf{x}'}_\mathcal{W}$ is illustrated in \reffig{fig:orthogonal_projection}.
Vector from the origin of the cone $\mathbf{o}_\mathcal{W}$ to the given point $\mathbf{x}_\mathcal{W}$ is denoted as $\mathbf{\hat{u}} = (\mathbf{x} - \mathbf{o})/\|\mathbf{x} - \mathbf{o}\|$.
Firstly, we find the angle between the cone direction $\mathbf{\hat{d}}$ and $\mathbf{\hat{u}}$ as $\alpha = \angle(\mathbf{\hat{u}}, \mathbf{\hat{d}})$.
Then, the angle between the cone surface and the desired orthogonal projection of $\mathbf{x}$ to the cone surface is $\beta = \alpha - \Theta$.
To find the projection, we first find $\mathbf{x}_r$ by rotating $\mathbf{x}$ around the vector $\mathbf{\hat{d}} \times \mathbf{\hat{u}}$ by the angle $\beta$.
The vector towards the projection is then constructed as $\mathbf{\hat{v}} = (\mathbf{x}_r - \mathbf{o})/\|\mathbf{x}_r - \mathbf{o}\|$.
Finally, the orthogonal projection of $\mathbf{x}$ on the cone surface is:
\begin{equation}
  \mathbf{x}' = \begin{cases}
    \mathbf{o} + \mathbf{\hat{v}}\cos\beta, & \text{if}\ \alpha < \pi/2,\\
    \mathbf{o}, & \text{if}\ \pi/2 \leq \alpha.
  \end{cases}
\end{equation}

%%{ Fig: Orthogonal projection

%%}

%%}

%%{ Sub: LKF

\subsection{Linear Kalman Filter}

\acp{LKF} filter uses a model with the hypothesis $\mathbf{x}_{[t]}$ and hypothesis covariance $\mathbf{\Omega}_{[t]}$:
\begin{equation}
  \mathbf{x}_{[t]} = \m{A}\mathbf{x}_{[t-1]},~\mathbf{A} = \left[\begin{smallmatrix}
    1 & 0 & 0\\
    0 & 1 & 0\\
    0 & 0 & 1
  \end{smallmatrix}\right],
  \label{eq:lti_model}
\end{equation}
where $\mathbf{A}$ is designed for a stationary 3D target.
Measurement noise covariance with the correction being applied in the direction of the $\mathbf{\hat{e}}_3$ axis is defined as
\begin{equation}
  \mathbf{R_e} = \left[\begin{smallmatrix}
    r & 0 & 0 \\
    0 & 10^9 & 0 \\
    0 & 0 & 10^9
  \end{smallmatrix}\right],
\end{equation}
where $r$ is used to weigh the correction in the direction to the cone surface.
The rest of the matrix is designed to mitigate corrections for other directions.
This canonical covariance $\mathbf{R}_e$ is then rotated every time to align with the vector $\mathbf{x}' - \mathbf{x}$ as $\mathbf{R} = \mathbf{P}\mathbf{R_e}\mathbf{P}^{\intercal}$
where $\mathbf{P}$ is a rotation matrix corresponding to rotation around the axis $\mathbf{\hat{e}}_3 \times \left(\mathbf{x}' - \mathbf{x}\right)$ by the angle $\angle \left(\mathbf{x}' - \mathbf{x}, \mathbf{\hat{e}}_1\right)$.
An obtained measurement is then the projection $\mathbf{x}'$ and the covariance $\mathbf{R}$.

Two variants of the filter were tested:
the fully-3D approach presented above, and a simplified 2D approach in which we assume that the radiation source is located on the ground.
In the latter case, we project the hypothesis onto the ground plane after each correction.

Initialization of the \ac{LKF} hypothesis cannot be done using the first arriving measurement as typical.
To tackle this issue, we first gather $N$ cone surfaces with different points of origin.
The gathering can be done using one of many exploration schemes, e.g., a traditional gradient-based exploration using particle intensity measurements as heuristics or flying a pre-determined sweeping path.
With the first $N$ cones measured, we initialize the filter by optimizing for a point $\mathbf{p}$, which has the minimum square distance to the cones' surfaces.
Firstly, we must define the distance to a cone surface.

%%}

%%{ Sub: Distance to a cone surface

\subsection{Distance to a cone surface}

Given a point $\mathbf{p}$ such that $\text{if}\ \mathbf{\hat{d}}^{\intercal}_{i}\mathbf{p} < \mathbf{\hat{d}}_i^{\intercal}\mathbf{o}_i$ ($\mathbf{p}$ is situated \emph{behind} the cone origin), the distance to the cone surface is trivial:
\begin{equation}
  d(\mathbf{p}, \mathbf{o}, \mathbf{\hat{d}}, \Theta) = \|\mathbf{o} - \mathbf{p}\|.
\end{equation}
Otherwise, the distance is expressed as
\begin{equation}
  d(\mathbf{p}, \mathbf{o}, \mathbf{\hat{d}}, \Theta) = \|\mathbf{p} - \mathbf{o}\| \sin \left( \cos^{-1}\frac{\left(\mathbf{p} - \mathbf{o}\right) \circ \mathbf{\hat{d}}}{\|\mathbf{p} - \mathbf{o}\|} - \Theta\right).
\end{equation}

%%}

%%{ Sub: Finding the closest point to multiple cone surfaces

\subsection{Finding the closest point to multiple cone surfaces}

The state estimator's initialization is done by finding the closest point to the first \emph{N} cone surfaces.
The point $\mathbf{p}_\mathcal{W}$ is found by minimizing the squared distance to each cone surface while satisfying geometric constraints:
\begin{align*}
  & \min_{\mathbf{p} \in \mathbb{R}^3} \label{eq:optimization_cost}\numberthis
  & \sum_{i=1}^{N}d(\mathbf{p}, \mathbf{o}_i, \mathbf{\hat{d}}_i, \Theta_i)^2
\end{align*}\begin{align*}
  \text{s.t.}~ \mathbf{\hat{d}}^{\intercal}_{i}\mathbf{p} &\geq \mathbf{\hat{d}}_i^{\intercal}\mathbf{o}_i, &\forall i &\in \{1, \hdots, n\}\numberthis\label{eq:constraint_cone_halfspace},\\
  \mathbf{p}^\intercal\mathbf{\hat{e}}_3 &= 0, &\forall i &\in \{1, \hdots, n\}\numberthis\label{eq:constraint_ground_plane}.
\end{align*}
The constraint (\ref{eq:constraint_ground_plane}) is used optionally in the case of the 2D estimator to enforce the search for a source near the ground plane.

This optimization problem is a nonlinear least-squares problem, which makes finding a near-optimum solution impractical and time-consuming.
However, we do not need to solve this problem quickly or in fast succession, and the solution is used only for initialization.
Therefore, we utilize the Sequential Least-Squares Programming optimization method.
The Jacobian of the criterion is calculated analytically.
The problem takes approx. \SI{1}{\second} to solve for $N = 5$ on an average onboard computer.

%%}

%%}

%%{ Experiments

\section{EXPERIMENTS}

The approach was verified in realistic simulations using the Gazebo/ROS simulator.
The Compton camera is modeled by combining real-time Monte-Carlo ray tracing \cite{baca2019timepix}.
Randomized simulations suggest that the proposed method is capable of initializing and further estimating the 3D position of the radiation source with no prior knowledge.
However, the \ac{UAV} should conduct motion orthogonally to the direction of the source.
Otherwise, both the \ac{LKF} initialization and the \ac{LKF} correction methods result in singular solutions that fail to provide a full 3D position of the source.
Similarly, the estimator fails when the \ac{UAV} is stationary.
This can be explained using the principles of the proposed method.
When the \ac{UAV} is stationary, all the cones originate from the same point.
Therefore, the information extracted from multiple cones only provide data regarding the direction to the source.
When the \ac{UAV} conducts a motion directly towards the radiation source, again, the information extracted from the cones do not provide us with other than direction data.
However, if the \ac{UAV} moves orthogonally to the direction to the source, a non-trivial intersection of multiple cones can exist (if they originate from different 3D positions) and therefore, the 3D position of the source can be estimated.
We can estimate the position of the source even in situations where traditional gradient-based method might fail, e.g., in urban environment.
The proposed method could be used as an addition to the intensity-based methods, that can utilize the same Timepix3 detector.
The orthogonal motion can be realized, e.g.,  by encircling the estimated hypothesis.

The presented two practical tests were conducted in cooperation of \emph{Advacam} (developer of the MiniPIX TPX3 detector), the CTU (developer of the proposed method and the aerial system) and the Czech Metrology Institute (provider of the sample radiation sources).
Only a few experiments were conducted due to the autumn/winter conditions in our country and the current social distancing rules, that have made such real-world tests increasingly difficult to organize.

\subsection{Simulations}

\begin{figure}[!ht]
  \centering
  \subfloat {
    \includegraphics[width=0.45\textwidth]{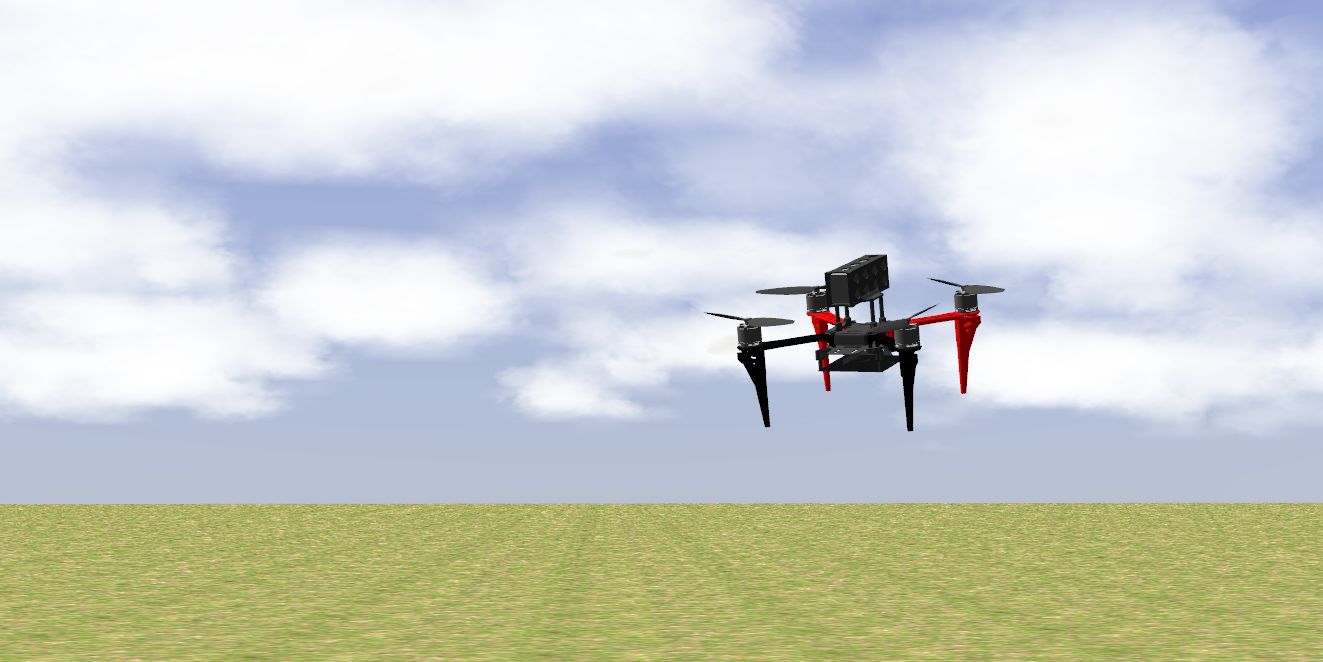}
  }
  \hfill%
  \caption{The simulation model of the DJI F330 UAV.}
  \label{fig:simulated_f330}
\end{figure}

%%{ Fig: simulation

\begin{figure*}[!ht]
  \centering
  \subfloat {
    \begin{tikzpicture}
      \node[anchor=south west,inner sep=0] (a) at (0,0) { \includegraphics[width=0.31\textwidth]{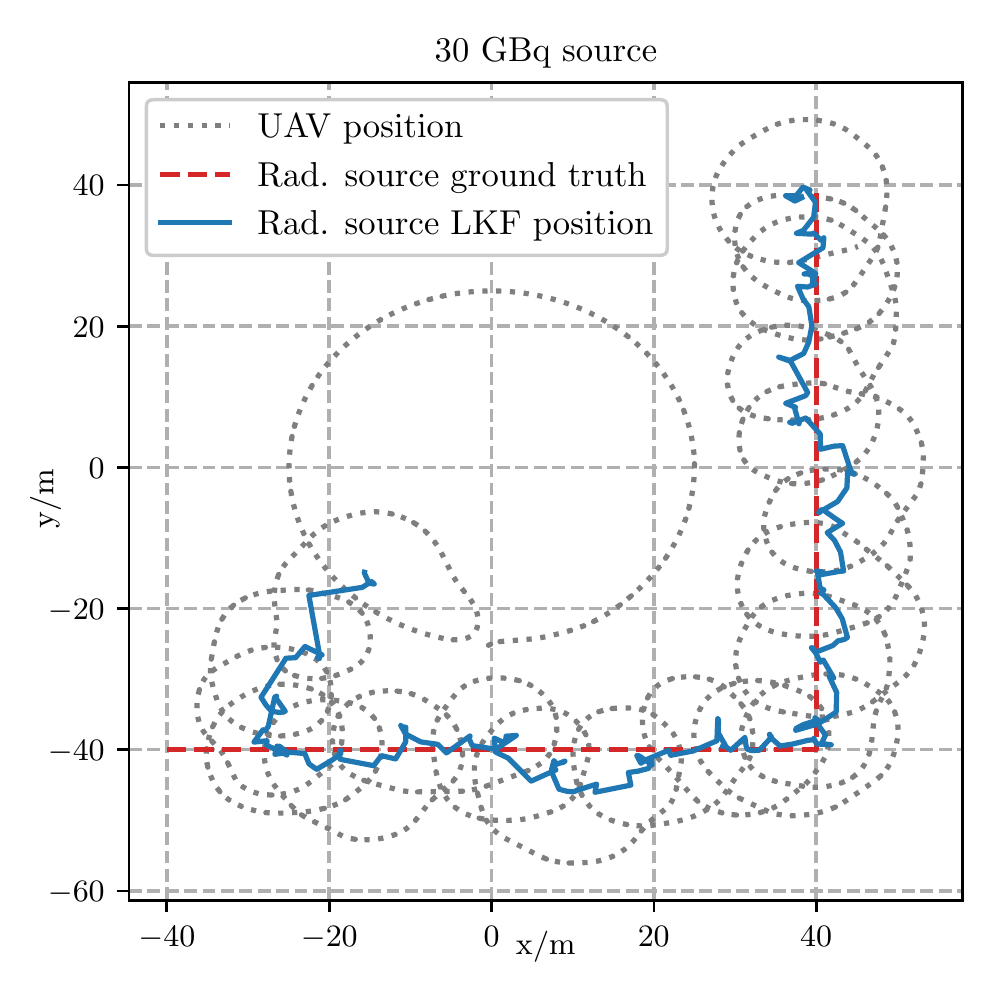}};
      \begin{scope}[x={(a.south east)},y={(a.north west)}]
        % \node[imgletter] (letter) at (0.0, 0.0) {(a)};
        % \draw (0.0, 0.0) rectangle (1.0, 1.0);
      \end{scope}
    \end{tikzpicture}
  }
  \subfloat {
    \begin{tikzpicture}
      \node[anchor=south west,inner sep=0] (a) at (0,0) { \includegraphics[width=0.31\textwidth]{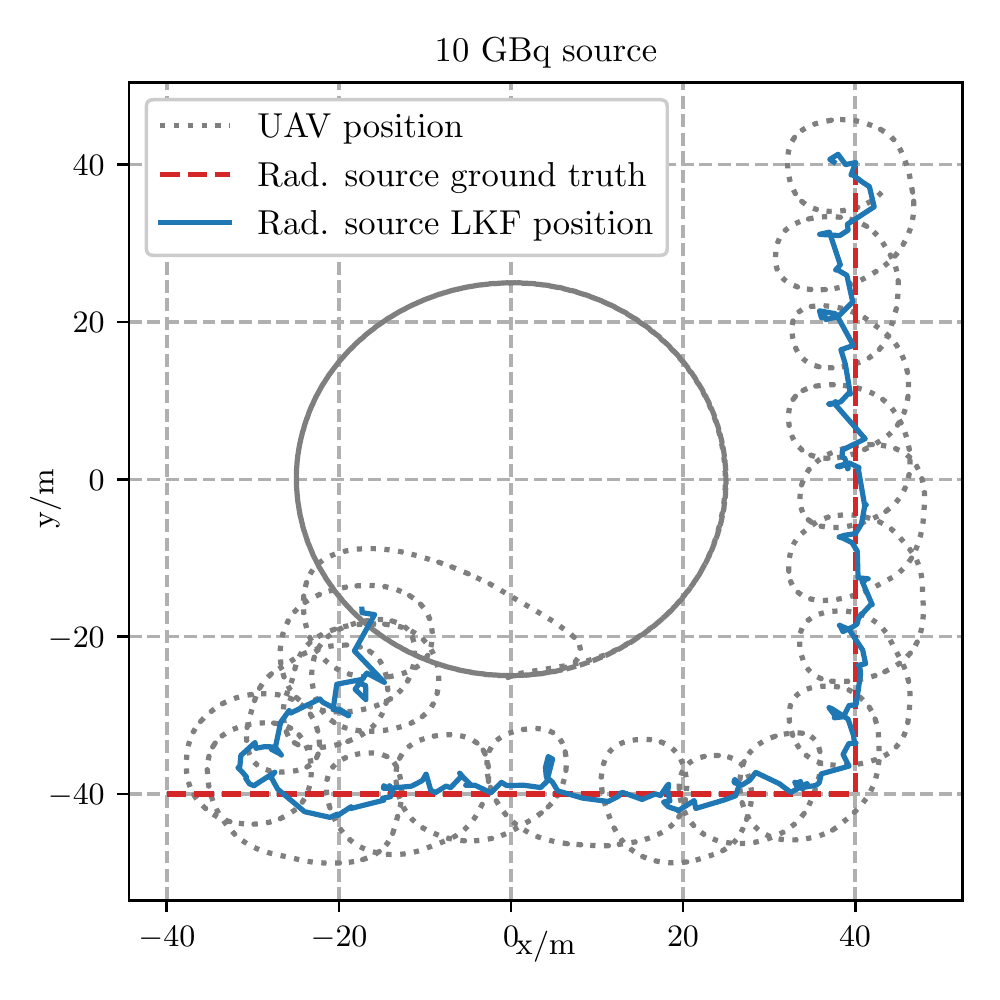}};
      \begin{scope}[x={(a.south east)},y={(a.north west)}]
        % \node[imgletter] (letter) at (0.0, 0.0) {(b)};
        % \draw (0.0, 0.0) rectangle (1.0, 1.0);
      \end{scope}
    \end{tikzpicture}
  }
  \subfloat {
    \begin{tikzpicture}
      \node[anchor=south west,inner sep=0] (a) at (0,0) { \includegraphics[width=0.31\textwidth]{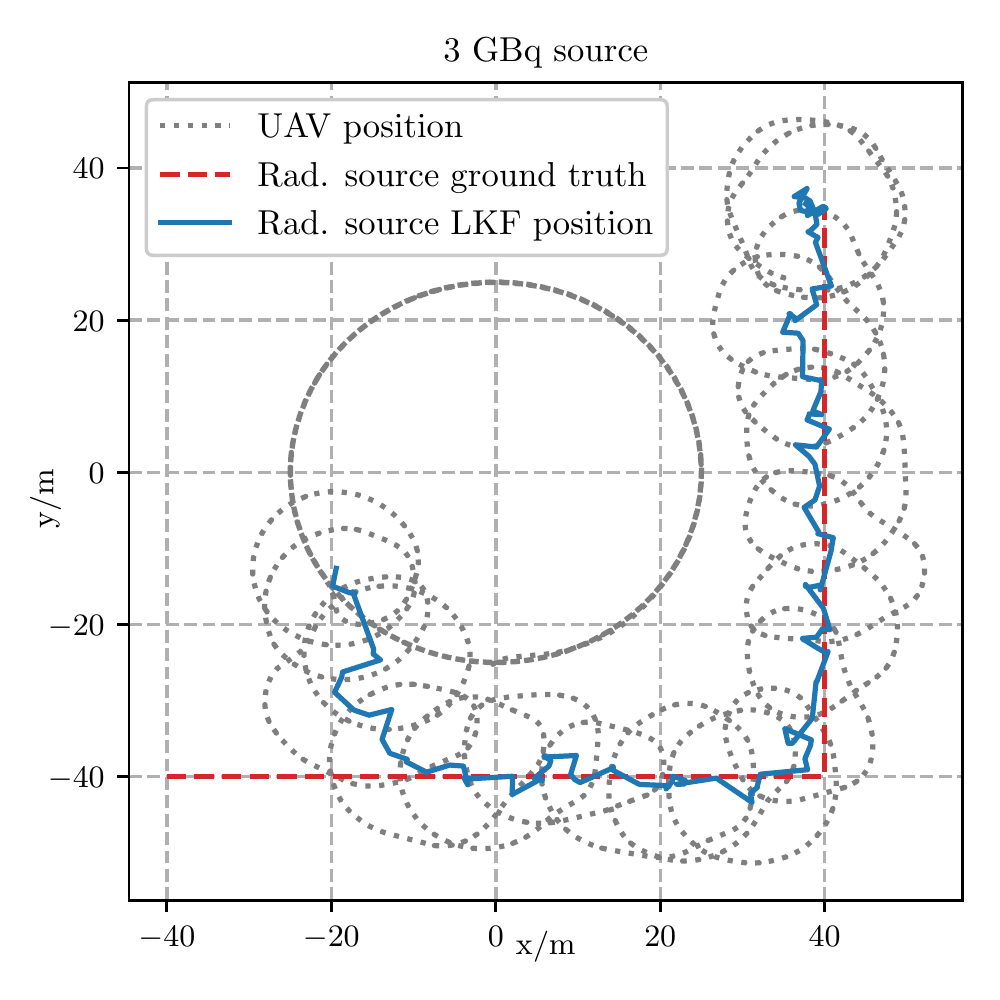}};
      \begin{scope}[x={(a.south east)},y={(a.north west)}]
        % \node[imgletter] (letter) at (0.0, 0.0) {(c)};
        % \draw (0.0, 0.0) rectangle (1.0, 1.0);
      \end{scope}
    \end{tikzpicture}
  }
  \caption{
    Top-down view of samples of simulations with a single radiation source, moving at the speed of \SI{1}{\meter\per\second}.
    First, the \ac{UAV} searched for the source by encircling the center of the designated $100\times100$\,\si{\meter} area.
    When the source for localized, the \ac{UAV} then proceeded with encircling the 2D \ac{LKF} hypothesis by circles with \SI{10}{\meter} radius.
    The method showed good tracking performance for all tested source energies: \SI{30}{\giga\becquerel}, \SI{10}{\giga\becquerel}, \SI{3}{\giga\becquerel}, however, the initial search phase took increasingly longer with a weaker source.
  }
  \label{fig:simulations}
\end{figure*}

%%}

A real-time ray tracing simulation model proposed previously in \cite{baca2019timepix} was used to evaluate the proposed localization method in conjunction with the realistic Gazebo/ROS robotics simulator.
The DJI F330 quadrotor was modeled (see \reffig{fig:simulated_f330}) and provided as a part of the MRS UAV system \cite{baca2021mrs}.
Development of a research strategy is not in the scope of this paper, hence we evaluate the localization method using the following naive search strategy:
\begin{enumerate}
  \item the \ac{UAV} encircles a designated area until a hypothesis is initialized by Eq.\,(\ref{eq:optimization_cost}),
  \item the \ac{UAV} then proceeds to encircle the hypothesis, updating the \ac{LKF} with the incoming measurements,
  \item if more than 3 cones in a row exhibit an outlier condition, the hypothesis is reset using Eq.\,(\ref{eq:optimization_cost}).
\end{enumerate}
Simulations show that the proposed method is able to localize a radiation source even with the naive search strategy.
Moreover, the proposed localization method is even suitable for real time tracking of a moving \isotope[137]{Cs} radiation source.
As shown in \reffig{fig:simulations}, the \ac{UAV} is able to track a radiation source moving at the speed of \SI{1}{\meter\per\second}.
In the case of the \SI{3}{\giga\becquerel} source, the average rate of detected Compton cones was \SI{1.7}{\per\second}.
In can be observed that a steady tracking is achieved even with the \SI{3}{\giga\becquerel} source, despite a lack of a specialized model in the LKF for a moving target.
The simulation software together with an example scenarios is provided as open source at \url{https://github.com/rospix}.

\subsection{Unmanned aerial vehicles}

%%{ Fig: experiment

\begin{figure*}[!ht]
  \centering
  \subfloat {
    \begin{tikzpicture}
      \node[anchor=south west,inner sep=0] (a) at (0,0) { \includegraphics[width=0.31\textwidth]{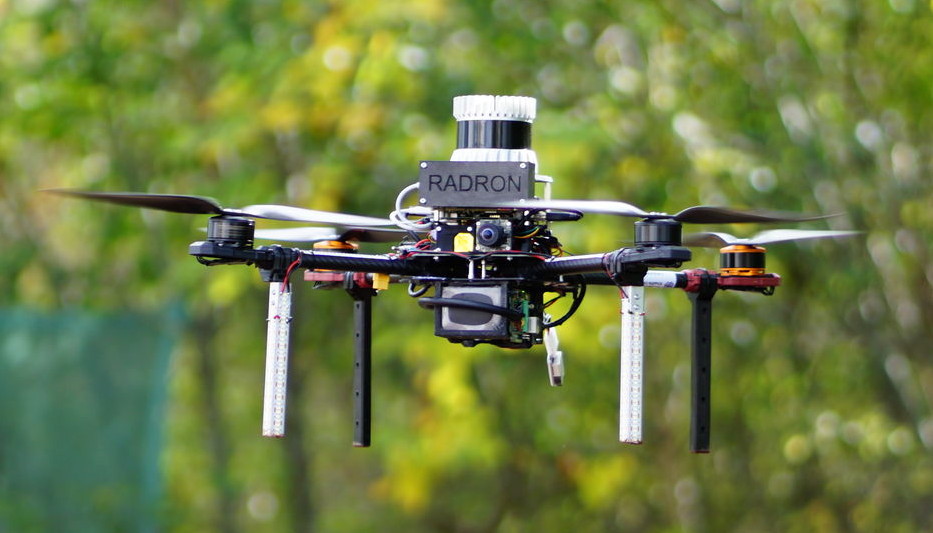}};
      \begin{scope}[x={(a.south east)},y={(a.north west)}]
        \node[imgletter] (letter) at (0.0, 0.0) {(a)};
        \draw (0.0, 0.0) rectangle (1.0, 1.0);
      \end{scope}
    \end{tikzpicture}
  }
  \subfloat {
    \begin{tikzpicture}
      \node[anchor=south west,inner sep=0] (a) at (0,0) { \includegraphics[width=0.31\textwidth]{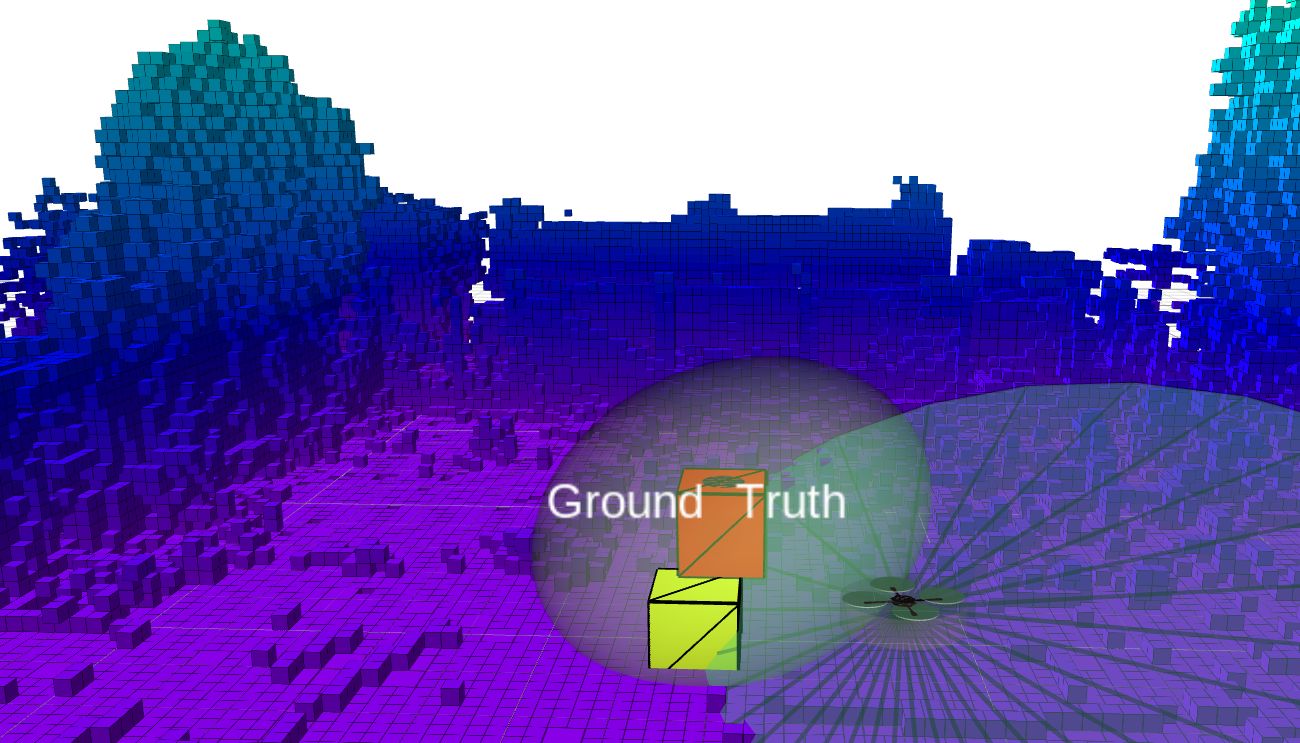}};
      \begin{scope}[x={(a.south east)},y={(a.north west)}]
        \node[imgletter] (letter) at (0.0, 0.0) {(b)};
        \draw (0.0, 0.0) rectangle (1.0, 1.0);
      \end{scope}
    \end{tikzpicture}
  }
  \subfloat {
    \begin{tikzpicture}
      \node[anchor=south west,inner sep=0] (a) at (0,0) { \includegraphics[width=0.31\textwidth]{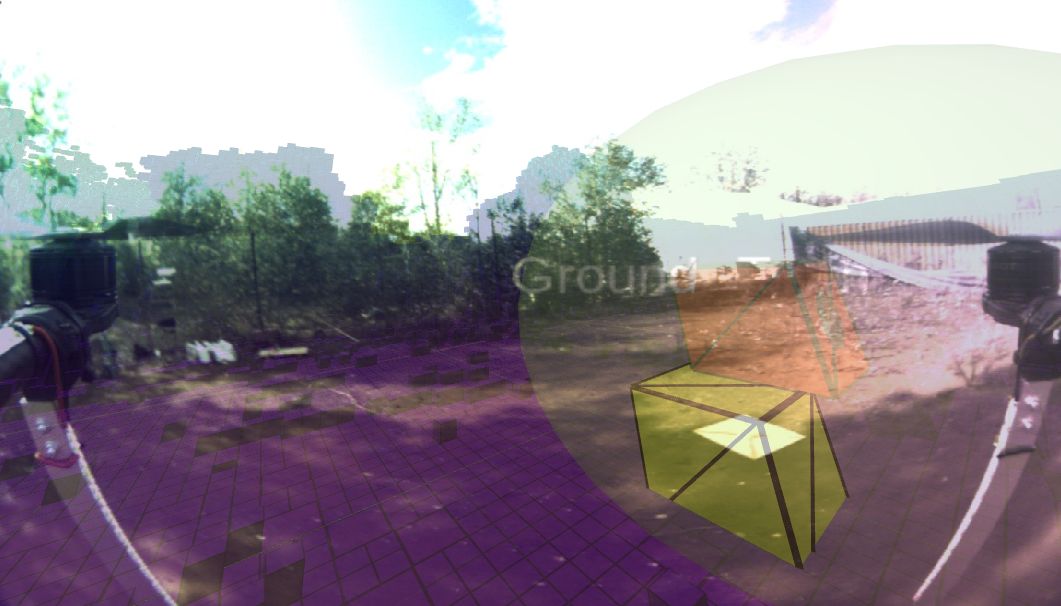}};
      \begin{scope}[x={(a.south east)},y={(a.north west)}]
        \node[imgletter] (letter) at (0.0, 0.0) {(c)};
        \draw (0.0, 0.0) rectangle (1.0, 1.0);
      \end{scope}
    \end{tikzpicture}
  }
  \caption{
    The first experimental flight: the T650 multi-rotor \ac{UAV} (a) was equipped with the MiniPIX TPX3 CdTe Compton camera, a 3D \ac{LiDAR}, and an RGB camera.
    \ac{SLAM} localizes the \ac{UAV} in 3D, which allows it to build a map of the surrounding environment (b).
    The RGB camera (c) serves for visual re-projection of the source's position.
    The estimated position of the source is shown in comparison with the ground truth position.
    Videos from the experiment are available at \url{http://mrs.felk.cvut.cz/radron-icuas}.
  }
  \label{fig:experiment}
\end{figure*}

%%}

The first verification of the MiniPIX detector onboard a \ac{UAV} was conducted using a general-purpose quadrotor helicopter built upon the Tarot 650 frame and equipped with Intel NUC-i7 onboard computer.
Figure \ref{fig:experiment}a depicts the \ac{UAV} in mid-air.
The Ouster OS1-16 3D \ac{LiDAR} was used for ground truth localization and mapping together with the A-LOAM \ac{SLAM} \cite{zhang2014loam}.
Furthermore, the \ac{UAV} was also equipped with a wide-angle RGB camera to aid with data visualization.
These sensors are not required for the proposed approach.
The only requirement is the access to real-time 3D position and orientation state estimate of the \ac{UAV}.
Such a state estimate is a requirement for any autonomous mission.
It can be accommodated by, e.g., a \ac{GNSS} receiver coupled with an \ac{IMU} and a magnetometer, or by a visual-inertial navigation system such as \cite{qin2018vins}.
Therefore, the Compton camera system can be employed on much smaller \ac{UAV} thanks to its small weight and size.
The design and size of the \ac{UAV} body is more limited by the localization sensor suite rather the by the \SI{40}{\gram} radiation detector.
The \ac{UAV} utilized the open-source \emph{MRS UAV system}\footnote{\url{http://github.com/ctu-mrs/mrs_uav_system}}\cite{baca2021mrs} in \ac{ROS} for automatic control and state estimation.

In the pursuit of miniaturization, a second experiment was conducted using the DJI F330 UAV (see Fig.~\ref{fig:experiment2}a), weighting approx. \SI{1}{\kilo\gram}.
The UAV also conducted an autonomous flight, however, this time using a monocular visual \ac{SLAM}, the VINS mono \cite{qin2018vins}.
A wide-angle greyscale camera was used for localization in real-time.

\subsection{Compton camera}

The Compton camera is built upon the Timepix3 detector~\cite{poikela2014timepix3} with a \SI{2}{\milli\meter} \ac{CdTe} sensor.
\ac{CdTe} sensors are exceptionally sensitive to high-energy gamma rays.
The sensitivity to high-energy $\gamma$ outweighs the small size (therefore small effective volume) of the sensor since only high-energy $\gamma$ can travel hundreds of meters through the air without being absorbed.
This allows sensing the sources from a further distance than using traditional detectors.
The Timepix3 detector with thick \ac{CdTe} sensor can detect photons ranging from 5 to \SI{1000}{\kilo\electronvolt} with relatively high quantum efficiency especially for Compton scattering events.
The miniaturized Timepix3-based detector MiniPIX TPX3 \ac{CdTe} with incorporated data acquisition, calibration and event reconstruction software layers was developed by Advacam (see \reffig{fig:timepix}).
The device does not require any cooling and can operate at room temperature.
The Compton camera payload weighs \SI{40}{\gram} and has dimensions of $80 \times 20 \times 15$\,\si{\milli\meter}.
The processing software is implemented in \ac{ROS} and is available as open-source.

\subsection{Experimental environment}

These preliminary tests were conducted in $40 \times 20$\,\si{\meter} area with a single \isotope[137]{Cs} source with activity of \SI{187}{\mega\becquerel} ($187 \times 10^{6}$ decay events per second).
This was a relatively weak source, sending only $\approx\,$ \SI{30}{\per\second\per\centi\meter\squared} photons at \SI{10}{\meter} distance.

The first real-world test\footnote{The first test video: \url{https://youtu.be/oH4jMMHfGVA}} was conducted under autonomous localization, mapping, and control, using the Tarot 650 UAV.
A human operator provided the UAV with control waypoints, and the \ac{UAV} followed them with \SI{1}{\meter\per\second} speed.
The \ac{UAV} first conducted a sweeping flight while gathering the first 5 cone surfaces for initialization.
After the \ac{LKF} was initialized, it started encircling the current hypothesis.
Such a strategy shows success in simulations.
After successfully localizing the source, the estimator can adapt to the potentially changing position of the source.
Figure \ref{fig:experiment} depicts the 3D map of the environment with the hypothesis showed as a red cube surrounded by a covariance ellipsoid.
A yellow cube marks the ground truth position of the source.

\begin{figure*}[!ht]
  \centering
  \subfloat {
    \begin{tikzpicture}
      \node[anchor=south west,inner sep=0] (a) at (0,0) { \includegraphics[width=0.31\textwidth]{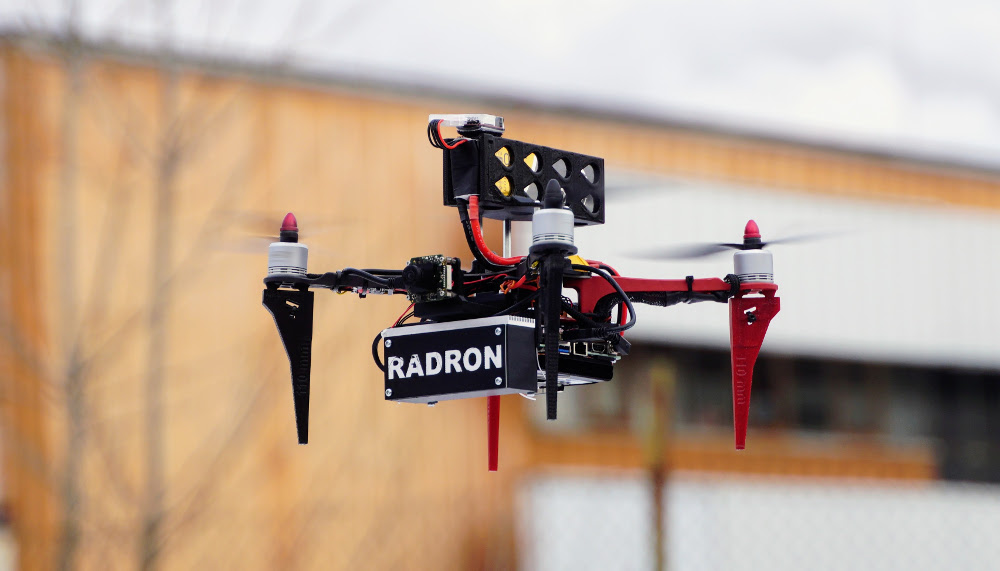}};
      \begin{scope}[x={(a.south east)},y={(a.north west)}]
        \node[imgletter] (letter) at (0.0, 0.0) {(a)};
        \draw (0.0, 0.0) rectangle (1.0, 1.0);
      \end{scope}
    \end{tikzpicture}
  }
  \subfloat {
    \begin{tikzpicture}
      \node[anchor=south west,inner sep=0] (a) at (0,0) { \includegraphics[width=0.31\textwidth]{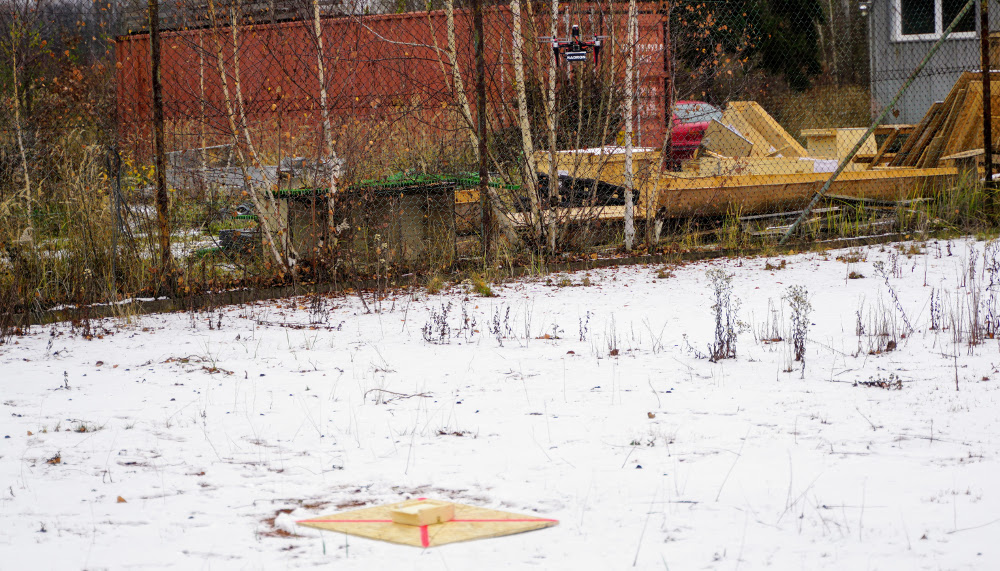}};
      \begin{scope}[x={(a.south east)},y={(a.north west)}]
        \node[imgletter] (letter) at (0.0, 0.0) {(b)};
        \draw (0.0, 0.0) rectangle (1.0, 1.0);
        \draw [white, ultra thick] (0.575, 0.91) circle (0.08);
      \end{scope}
    \end{tikzpicture}
  }
  \subfloat {
    \begin{tikzpicture}
      \node[anchor=south west,inner sep=0] (a) at (0,0) { \includegraphics[width=0.31\textwidth]{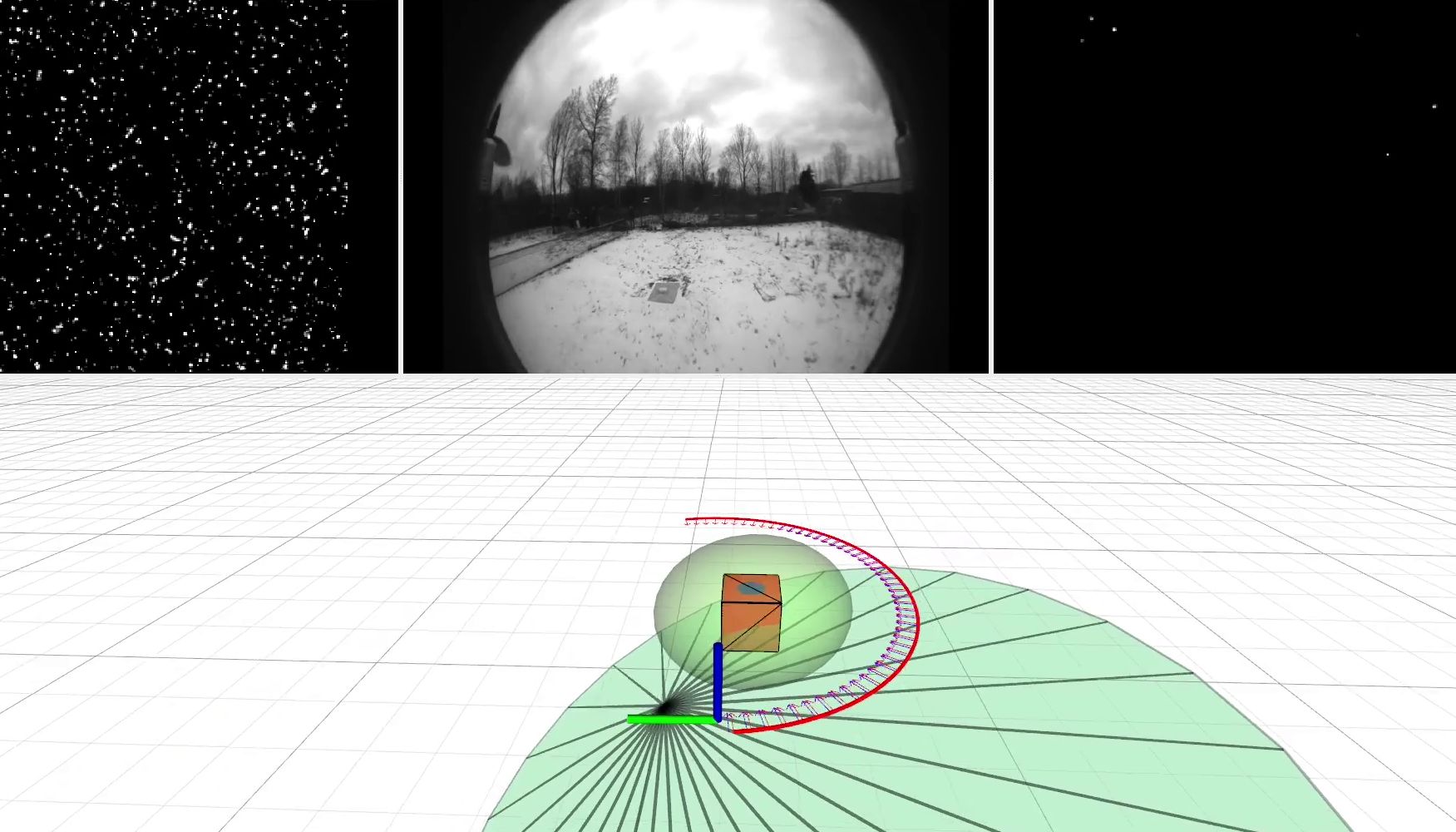}};
      \begin{scope}[x={(a.south east)},y={(a.north west)}]
        \node[imgletter] (letter) at (0.0, 0.0) {(c)};
        \draw (0.0, 0.0) rectangle (1.0, 1.0);
      \end{scope}
    \end{tikzpicture}
  }
  \caption{
    The second experimental flight: the DJI F330 multi-rotor \ac{UAV} (a), (b) was equipped with the MiniPIX TPX3 CdTe Compton camera, and a monocular grey-scale camera.
    VINS MONO \ac{SLAM} was used to provide the \ac{UAV} with 3D odometry.
    A snapshot from the video \url{http://mrs.felk.cvut.cz/radron-icuas} is shown, depicting a 3D plot of the UAV's position, the latest Compton cone and the estimated position of the radiation source.
  }
  \label{fig:experiment2}
\end{figure*}
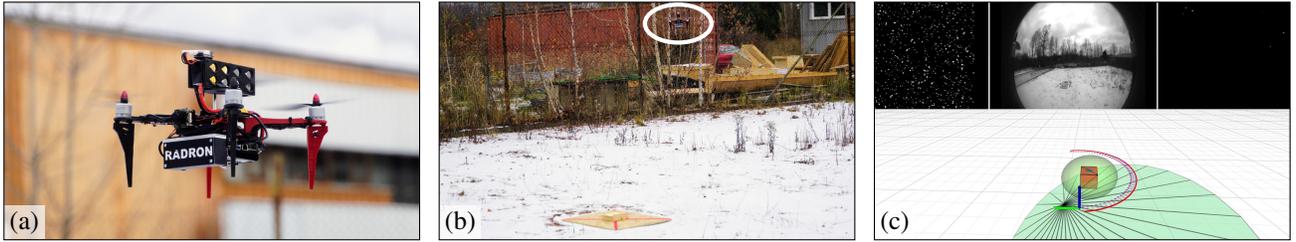

The real-world test\footnote{The second test video: \url{https://youtu.be/t6ayhHJ9z5k}} was conducted using the miniaturized UAV, the DJI F330.
Again, the flight was conducted autonomously.
This time, the point of the experiment was to test an updated version of the MiniPIX TPX3 detector API, developed specifically for detecting the Compton coincidences within the embedded hardware of the detector.
Therefore, the UAV was tasked to autonomously encircle the already known position of the radiation source.
Figure \ref{fig:experiment2} depicts the \ac{UAV} in mid-air, together with a snapshot of a 3D visualization of the recorded data.

Full-flight data from the experiments, including the RAW sensor data, localization data, LiDAR point clouds and camera images are available on the website \url{http://mrs.felk.cvut.cz/radron-icuas} in the form of \emph{rosbag} files.

\subsection{Results}

%%{ Fig: results

\begin{figure*}[!ht]
  \centering
  \includegraphics[width=1.0\textwidth]{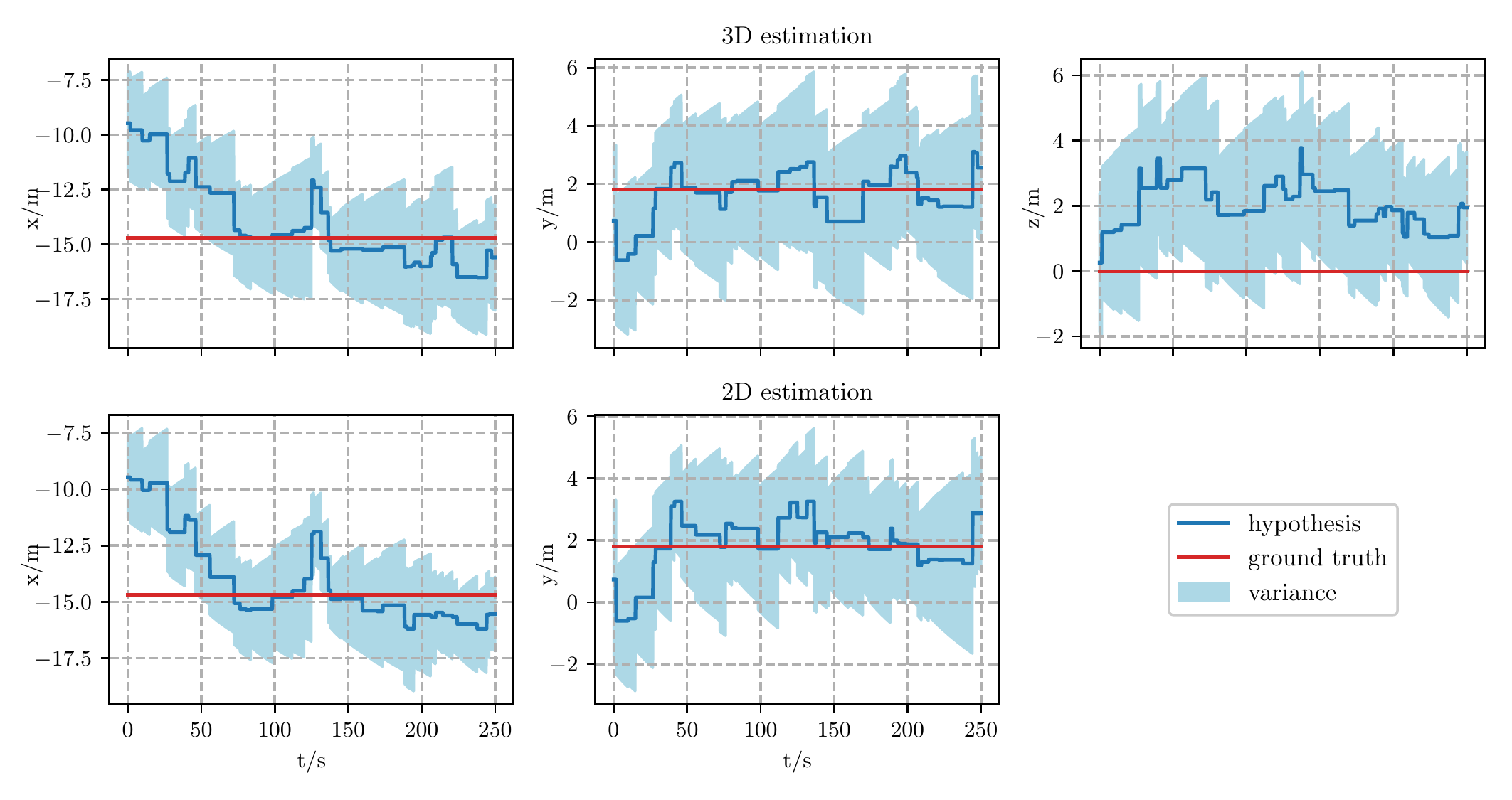}
  \caption{Plots of convergence of the 3D hypothesis and the 2D hypothesis in comparison to the ground truth. Note that the vehicle maintained a nearly constant height above ground during the experiment, hence the slow convergence in the Z-axis.}
  % \vspace{-1em}
  \label{fig:results}
\end{figure*}

%%}

Statistics of event occurrences observed during the experiments is shown in Table~\ref{tab:results}.
Figure \ref{fig:results} shows plots of the estimated 3D and 2D position of the gamma source in comparison to the ground truth.
The hypothesis was correctly estimated near the ground truth after gathering just 15 Compton events.
The experimental data show the estimator system working as expected.
Future work should emphasize on rejecting outliers.
Outliers can occur due to the omnipresent radiation background, even though occurrence of coincidences in the radiation background is extraordinarily rare.
The detector registers only approx. 10 particles per second due to the radiation background, the chance of two background particles coinciding within the time frame of approx. \SI{100}{\nano\second} is extremely low.
The largest source of outliers lies in the uncertainty of origin of the already detected two coinciding particles.
In some cases their type (electron, scattered photon) can be deduced from the measured energies using, e.g., the knowledge of the Compton edge for a known isotope.
However, if both measured energies lie in the Compton continuum, the particle type is ambiguous.
Moreover, even if the incoming photon comes from a known direction, back-scattering can make the measured by-products appear as incoming from an \emph{opposite} direction.
However, despite the occasional outliers, the estimator performed well despite the tested radiation source's low activity.
It must be noted, that majority of the measured signal, that can be seen in the provided videos and in the provided datasets, is a valid measurement of the incoming gamma photons, not of a background radiation.
A majority of the signal originates from photons measured by pure photo-absorption, therefore, the signal does not produce coinciding particles used for Compton reconstruction.
Nevertheless, the signal could be used for traditional intensity-based estimation \cite{mascarich2018radiation}, since the CdTe Timepix3 detector is fully capable to serve as a dosimeter.
The method proposed in this manuscript aims to provide an additional information that could aid the traditional state-of-the-art methods, without the need for extra hardware.

%%{ Tab: results

\begin{table}[hb!]
  \centering
  \begin{tabular}{r|ccc}
    & Event count & Rate [s$^{-1}$] & Relative share \\
    \hline
      % Photoelectric & 13697 & 54.79 & 0.996 \\
      % Compton & 39 & 0.156 & 0.003 \\
      % E $>$ 800keV & 13 & 0.052 & 0.001 \\
      Photoelectric & 6073 & 24.29 & 0.996 \\
      Compton & 18 & 0.072 & 0.003 \\
      Background & 8 & 0.032 & 0.001 \\
    \hline
      Photoelectric & 5315 & 21.26 & 0.996 \\
      Compton & 18 & 0.072 & 0.003 \\
      Background & 4 & 0.016 & 0.001 \\
  \end{tabular}
  \caption{Events captured by the Compton camera during the first 250 seconds of a flight. Data from Experiment 1 are shown in the upper section, and Experiment 2 in the lower section. Events exceeding energy of \SI{800}{\kilo\electronvolt} are labeled as background, as these are too energetic to be released by the \isotope[137]{Cs} source.}
  \label{tab:results}
\end{table}

%%}

%%}

%%{ Fig: experiment2

%%}

%%{ Future work

  \section{LESSONS LEARNED AND FUTURE WORK}

  The presented results are a pilot work for a project for autonomous localization and tracking of compact ionizing radiation sources by \aclp{MAV}.
  Several significant challenges have been identified that are yet to be tackled before the solution becomes fully autonomous and reliable.
  The single-detector Compton camera should be compared to a multi-sensor stack.
  Although the proposed single-detector variant offers a significantly simpler hardware solution, the angular resolution is worse than with a sensor stack.
  Moreover, reconstruction of cone surfaces using (\ref{eq:theta_equation}) should benefit from outlier rejection using the values of $E_\lambda, E_{\lambda'}, E_{e^{-}}$.
  Priors of $E_\lambda$ can be estimated in real-time by analyzing the spectrum, a type of incoming particles.
  Also, the state estimation hypothesis should be generalized to a mixture of distributions.
  In combination with the observed complex 3D map of the environment, Monte-Carlo methods such as the particle filter should be explored.
  If available, prior knowledge of the environment structure, e.g., wall material and thickness, should also be utilized by the estimator.

  We pursue the use of \acp{MAV} equipped with monocular visual \ac{SLAM} (see \reffig{fig:timepix}).
  Autonomous planning methods should simultaneously explore an unknown environment while maximizing the yield of information gathered by the Compton camera.
  This task shares goals with the \ac{DARPA} SubT challenge\footnote{\url{https://www.subtchallenge.com}} in which the authors participate.
  Simulations show that the multi-\ac{MAV} system benefits significantly from sharing the measured data between the \acs{MAV} \cite{stibinger2020localization}.
  We aim to utilize swarms of \acp{MAV} to increase the \emph{baseline} of the distributed sensor, which increases the convergence speed of the estimator.

%%}

%%{ Conclusions

\section{CONCLUSIONS}

We have showcased a novel approach for localization and position estimation of a gamma radiation source using a miniature single-detector Compton camera on an \acl{MAV}.
The proposed method of estimating the source's position utilizes the particle physics interactions within the \ac{CdTe} sensor in conjunction with a \acl{LKF}.
A novel method of the filter initialization is proposed, given the Compton camera detector's unique form of measurements.
Moreover, a novel iterative-based correction method is proposed to fuse the measurements in the form of a cone surface.
The methods were validated by real-world experiments showing promising results that were, until now, the domain of heavy-lifting aircraft.
Gazebo/ROS simulation plugin for the Compton camera and the processing software is provided as open-source\footnote{\url{http://github.com/rospix}} to allow and motivate research on autonomous systems using the Compton camera detector.

%%}

%%{ Acknowledgment

\section*{ACKNOWLEDGMENT}

This work was done on behalf of Medipix3 collaboration, was supported by TACR project no. FW01010317, and by CTU grant no SGS20/174/OHK3/3T/13.

%%}

% \addtolength{\textheight}{-12cm}

% \clearpage

\bibliographystyle{IEEEtran}
\bibliography{main}

\end{document}